\documentclass[letterpaper, 10 pt, conference]{ieeeconf}
\IEEEoverridecommandlockouts 
\usepackage{graphics} 
\usepackage{amsmath} 
\usepackage{amssymb}  
\usepackage{cite}
\usepackage{amsfonts}
\usepackage{bm}
\usepackage{breqn}

\usepackage{diagbox}
\usepackage{multirow}
\usepackage{multicol}
\usepackage{graphicx}
\usepackage[subtle,tracking=normal]{savetrees}
\usepackage{balance}
\usepackage{caption}
\usepackage{subcaption}
\usepackage{makecell}
\usepackage{soul}

\usepackage{hyperref}
\usepackage[percent]{overpic}
\usepackage{xcolor}
\newcommand{\insertYoutubeLink}{\url{https://youtu.be/OLoWSX_R868}}

\title{\LARGE \bf
AllGaits: Learning All Quadruped Gaits and Transitions 
}

\author{Guillaume Bellegarda, Milad Shafiee, Auke Ijspeert%
\thanks{
This research is supported by the Swiss National Science Foundation
(SNSF) as part of project No.197237. The authors are with the BioRobotics Laboratory, Ecole Polytechnique Federale de Lausanne (EPFL).
 {\tt \{firstname.lastname\}@epfl.ch}}%
}

\begin{document}
\bstctlcite{MyBSTcontrol}
\maketitle
\thispagestyle{empty}
\pagestyle{empty}

\begin{abstract}
We present a framework for learning a single policy capable of producing all quadruped gaits and transitions. The framework consists of a policy trained with deep reinforcement learning (DRL) to modulate the parameters of a system of abstract oscillators (i.e.~Central Pattern Generator), whose output is mapped to joint commands through a pattern formation layer that sets the gait style, i.e.~body height, swing foot ground clearance height, and foot offset. Different gaits are formed by changing the coupling between different oscillators, which can be instantaneously selected at any velocity by a user. With this framework, we systematically investigate which gait should be used at which velocity, and when gait transitions should occur from a Cost of Transport (COT), i.e.~energy-efficiency, point of view. Additionally, we note how gait style changes as a function of locomotion speed for each gait to keep the most energy-efficient locomotion. While the currently most popular gait (trot) does not result in the lowest COT, we find that considering different co-dependent metrics such as mean base velocity and joint acceleration result in different `optimal' gaits than those that minimize COT.  We deploy our controller in various hardware experiments, showing all 9 typical quadruped animal gaits, and demonstrate generalizability to unseen gaits during training, and robustness to leg failures. Video results can be found at \insertYoutubeLink. 
\vspace{-0.5em}
\end{abstract}

\vspace{-0.2em}
\section{Introduction}
\vspace{-0.3em}
\label{sec:introduction}

Animals seamlessly transition between different gaits as they change speeds, or to react to variable terrain. Such transitions emerge through inter-limb coordination governed by the interaction between the brain, the spinal cord, and the musculoskeletal system~\cite{grillner2020current}. Several hypotheses have been proposed as explanations for why the gait transitions occur: to minimize energy expenditure~\cite{hoyt1981gait}, minimize peak musculoskeletal forces~\cite{farley1991mechanical}, maximize periodicity~\cite{granatosky2018inter}, and, recently, promote \textit{viability}, by formalizing the notion of avoiding a fall during locomotion~\cite{shafiee2024viability}. Gaits themselves can be formed through several different biological mechanisms, for example sensory driven~\cite{owaki2013simple}, through descending drive~\cite{harischandra2011sensory}, or coupling driven through Central Pattern Generators (CPGs) in the spinal cord~\cite{ijspeert2008central}.

In robotics, quadruped robots are displaying complex motor skills with different gaits to locomote at varying speeds and across challenging terrains, including combinations of discrete capabilities like running and jumping~\cite{miki2022learning,cheng2023parkour,zhuang2023robot,bellegarda2024robust,bellegarda2024quadruped,margolis2022pixels,smith2023learning}. While several works study transitions between such gaits (for example), the optimal transition times, speeds, and between which discrete gaits remains an open question. Additionally, for frameworks that do show transitions between gaits, the parameters must often be re-tuned for each (MPC)~\cite{dicarlo2018mpc}, may have heuristics for transitioning~\cite{shao2022gait}, or may otherwise be non-optimal, as the cyclic motions may affect the body and joints differently. For example, in contrast with most robots (with some exceptions~\cite{seok2013design,khoramshahi2013benefits}), animals do not bound with a rigid spine. 

\begin{figure}
\vspace{-0.3em}
    \centering
     \includegraphics[width=0.93\linewidth]{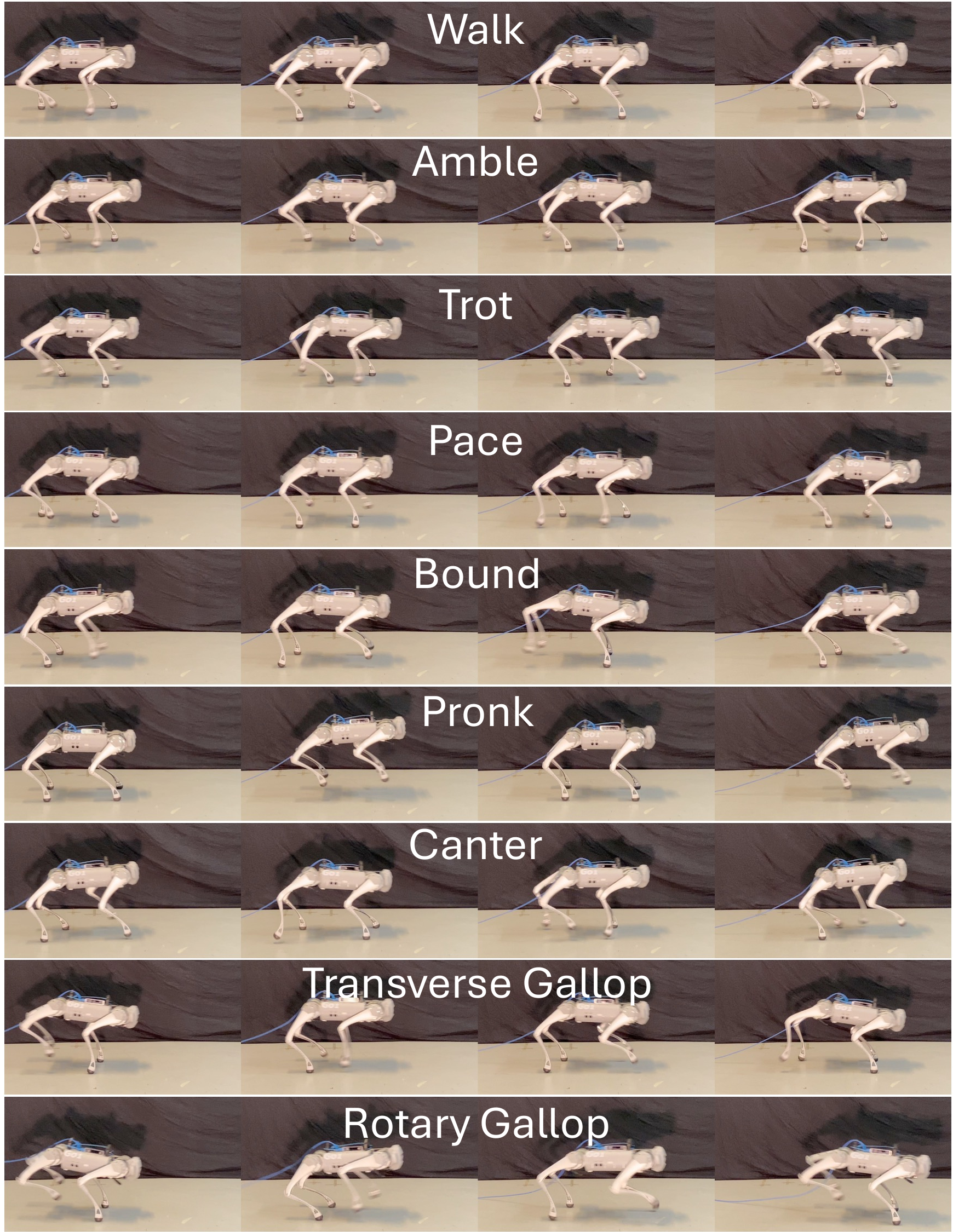}\\
     \vspace{-0.3em}
    \caption{\textbf{AllGaits}: snapshots from learning all quadruped gaits with Central Pattern Generators and deep reinforcement learning. 
    }
    \label{fig:intro}
    \vspace{-2.2em}
\end{figure}

For quadruped robots, gaits can be encoded in a biological neural network (CPG)~\cite{ijspeert2008central,sprowitz2013cheetah,righetti08,zhang2024online}, enforced through a fixed pattern in optimal control frameworks~\cite{dicarlo2018mpc,kim2019highly,bellicoso2018dynamic}, specifically rewarded in learning frameworks through a reward function~\cite{margolis2022walktheseways,wu2023learning}, or can emerge naturally during the training process~\cite{bellegarda2022robust,bellegardaIROS19TaskSpaceRL,fu2022energy}. One observation is that learned controllers working at a wide range of speeds seem to converge to trot gaits, especially when the reward function has terms penalizing non-stable motions, as the trot gait minimizes body angular velocities even at high speeds~\cite{ji2022concurrent,margolis2022rapid}. This is in contrast to high-speed gaits found in nature, where flexible spines exhibit significant bending movement while quadruped animals bound or gallop~\cite{alexander1988mammals}, or for Springboks and Thomson's gazelles exhibit stotting (or pronking) behaviors~\cite{fitzgibbon1988stotting}. 

Reinforcement learning has been applied to directly imitate animal motions such as pacing and trotting through training separate policies that reward tracking different mocap data~\cite{peng2020laikagoimitation}. Generating a library of reference trajectories and training a goal-conditioned policy to imitate them, and explicitly providing transition and coupling strength timing can also lead to executing gait transitions~\cite{shao2022gait}. 
By training three policies to locomote at specific velocities (0.375 \texttt{m/s}, 0.9 \texttt{m/s}, 1.5 \texttt{m/s}) while minimizing energy consumption, three corresponding distinct gaits emerged (walk, trot, pronk)~\cite{fu2022energy}. Then, using these three policies as experts, the transition between different speeds and gaits was realized to locomote between $0.375-1.5$ \texttt{m/s}. Transitions between gaits can also be realized by training a high-level gait policy that specifies gait patterns of each foot, while a low-level convex MPC controller optimizes the motor commands so that the robot can walk at a desired velocity using that gait pattern~\cite{yang2022fast}. 

Combining learning with bio-inspired representations of neural circuits allows for higher centers to modulate and coordinate gaits~\cite{bellegarda2022cpgrl,bellegarda2024visual,shafiee2024manyquadrupeds}, which can also result in the emergence of terrain-driven gait transitions to successfully cross variable gaps~\cite{shafiee2023puppeteer,shafiee2024viability}. As an alternative to learning-based control, quadruped robots have also demonstrated gait generation and transitions can occur through simple force feedback, without explicit coupling between oscillators~\cite{owaki2013simple,owaki2017quadruped}.

\vspace{-0.4em}
\subsection{Contribution} 
\vspace{-0.1em}
While gait transitions can occur through sensory feedback~\cite{owaki2013simple} and/or through descending drives~\cite{harischandra2011sensory}, in this paper we take a coupling-driven approach to learn to locomote with, and transition between, a variety of gaits. Coupling between different abstract oscillators is commonly used for CPG-based locomotion control of different gaits in bio-inspired robotics~\cite{ijspeert2008,sprowitz2013cheetah,righetti08}. 

While several common quadrupedal gaits have been successfully demonstrated on quadruped hardware, previous work requires either explicit parameter tuning in MPC~\cite{dicarlo2018mpc}, extensive reward function tuning~\cite{margolis2022walktheseways,shao2022gait}, specific training schemes~\cite{fu2022energy}, or expert demonstrations from animals or MPC to imitate~\cite{peng2020laikagoimitation}. In contrast, we show all quadruped gaits and their transitions can be realized without reward function tuning or any expert demonstrations. 

We center our scientific investigation around three fundamental biological and robotics locomotion questions: 
\begin{itemize}
    \item [1.] Which gaits are most efficient at which velocities, and when should gait transitions occur? 
    \item [2.] How should parameters like body height, posture, and swing foot trajectories change for different gaits at different velocities? 
    \item [3.] Can we produce novel gaits not seen during training, and how robust is the policy to leg failures? 
\end{itemize}

In order to answer these questions, we present a hierarchical bio-inspired architecture consisting of a policy trained with DRL (higher centers), a network of coupled oscillators mapped to task space foot trajectories (rhythm generator and pattern formation layers of the spinal cord), and sensory feedback from onboard sensors and internal CPG states (efference copy of the spinal cord). We explicitly enforce a gait through the coupling matrix, and the locomotion style through the pattern formation parameters (i.e.~body height, swing foot ground clearance, foot offsets). We leverage this architecture to produce all quadrupedal gaits, determine when the optimal transitions between gaits should occur, and with which locomotion style. Additionally, we are able to realize novel gaits that were not seen during training, and have not been previously shown, without any modifications directly in hardware experiments. Furthermore, our framework is robust to failures of either one or two disabled legs. 

The rest of this paper is organized as follows. In Section~\ref{sec:method} we present AllGaits, including our design choices and integration of Central Pattern Generators into the deep reinforcement learning framework to learn to locomote with all quadruped gaits. In Section~\ref{sec:result} we discuss results and analysis from learning our controller and sim-to-real transfers, and we give a brief conclusion in Section~\ref{sec:conclusion}.

\vspace{-0.3em}
\section{Learning Central Pattern Generators \\ for All Quadrupedal Gaits}
\label{sec:method}
\vspace{-0.1em}

\begin{figure*}[!t]
      \vspace{-0.5em}
      \centering
      \includegraphics[width=\linewidth]{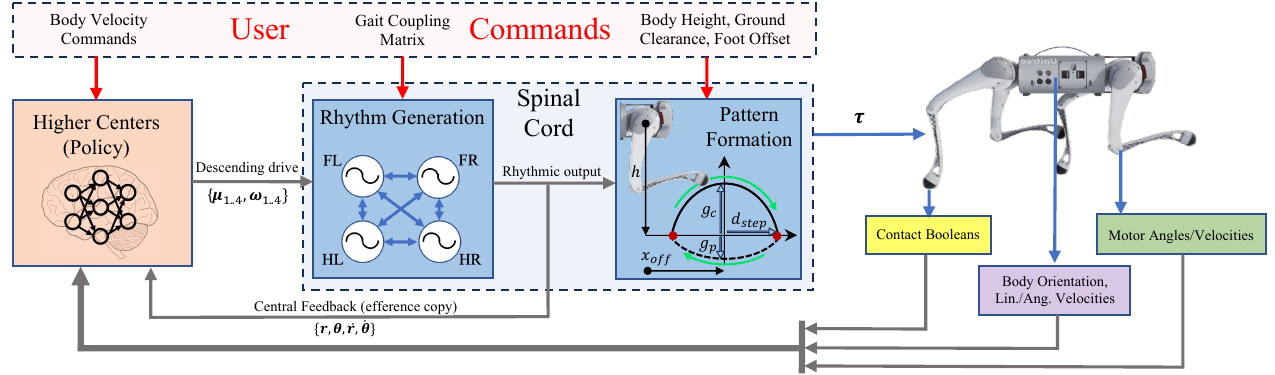}\\  
      \vspace{-0.4em}
      \caption{
       \textbf{AllGaits}: Control architecture for learning central pattern generators to locomote at all gaits for quadruped robots. The observation consists of velocity commands, proprioceptive measurements, and the current CPG states (efference copy of the spinal cord), which the policy network uses to select CPG parameters $\mu$ and $\omega$ for each leg $i$ (Front Right (FR), Front Left (FL), Hind Right (HR), Hind Left (HL)) to coordinate the Rhythm Generation. A gait coupling matrix is input from the user to set a particular gait. The resulting CPG states are then mapped to desired foot positions in a Pattern Formation layer, which the user can also directly modulate by setting body height $h$, swing foot ground clearance $g_c$, and foot offset from the hip $x_{off}$. This task-space mapping is then converted to desired joint angles with inverse kinematics, and finally tracked with joint PD control to produce torques $\bm{\tau}$. The control policy selects actions at 100 Hz, and all other blocks operate at 1 kHz.
      }
      \label{fig:control_diagram}
      \vspace{-1.3em}
\end{figure*}

\begin{figure*}
\centering
\vspace{0.5em}
\includegraphics[width=\linewidth]{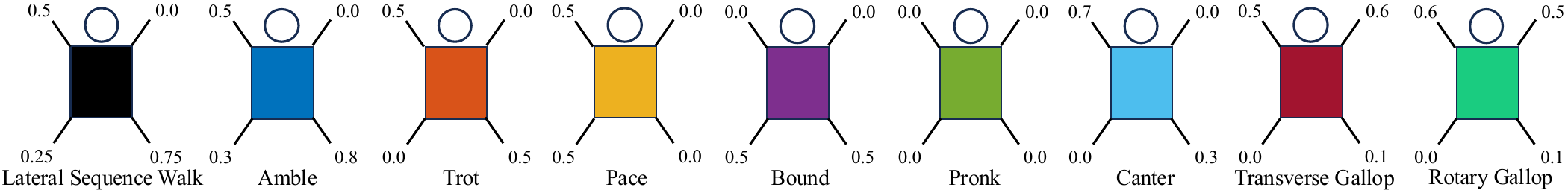}
\vspace{-1.4em}
\caption{Contact timing for each foot with the ground as a percentage of a single gait cycle for various quadruped gaits: Lateral Sequence Walk, Amble, Trot, Pace, Bound, Pronk, Canter, Transverse Gallop (T.G.), Rotary Gallop (R.G.). 
These timings are converted to matrices that denote phase offsets between different limbs in column order: Front Right (FR), Front Left (FL), Hind Right (HR), Hind Left (HL), as they appear in Equation~\ref{eq:salamander_theta}. 
}
\label{fig:gait_matrices}
\vspace{-2em}
\end{figure*}

In this section we describe our CPG-integrated reinforcement learning framework and design decisions for learning locomotion controllers to produce all quadrupedal gaits. Figure~\ref{fig:control_diagram} shows an overview of our framework and the parallel with biological systems, where output from higher centers (the policy network) modulates the spinal cord (rhythm generation and pattern formation layers), finally actuating the motors (muscles). We use CPG-RL~\cite{bellegarda2022cpgrl} as a basis, where now we include coupling matrices to define each of the following gaits: Walk, Amble, Trot, Pace, Bound, Pronk, Canter, Transverse Gallop, and  Rotary Gallop (Figure~\ref{fig:gait_matrices}). 

\vspace{-0.3em}
\subsection{Rhythm Generation and Pattern Formation}
\vspace{-0.1em}

The abstract oscillators which form our Rhythm Generation layer are defined as:
\vspace{-0.2em}
\begin{align}
\ddot{r}_i &= a \left(\frac{a}{4} \left(\mu_i - r_i \right) - \dot{r}_i \right) \label{eq:salamander_r} \\
\dot{\theta}_i &= \omega_i +\sum_{j}^{} r_j w_{ij} \sin(\theta_j - \theta_i - \phi_{ij}) \label{eq:salamander_theta}
\end{align}

\vspace{-0.5em}
\noindent where $r_i$ is the current amplitude of the oscillator, $\theta_i$ is the phase of the oscillator, $\mu_i$ and $\omega_i$ are the intrinsic amplitude and frequency, $a$ is a positive constant representing the convergence factor. Couplings between oscillators are defined by the weights $w_{ij}$ and phase biases $\phi_{ij}$.
For a quadruped robot with a single oscillator corresponding to each leg, the coupling matrices $\bm{\Phi}$  can be defined by following the timings from Figure~\ref{fig:gait_matrices}. The row/column order is  Front Right (FR), Front Left (FL), Hind Right (HR), Hind Left (HL). These matrices define the offsets between different oscillators to converge to the desired gaits. With appropriately high (strong) coupling weights, i.e.~$w_{ij}=10$, these coupling matrices enforce the gait. 

As in CPG-RL~\cite{bellegarda2022cpgrl}, to map from the oscillator states to joint commands, we first compute corresponding desired foot positions, and then calculate the desired joint positions with inverse kinematics. The desired foot position coordinates are given as follows:
\vspace{-0.7em}
\begin{align}
x_{i,\text{foot}} &= x_{off}-d_{step} (r_i-1) \cos(\theta_i) \\
z_{i,\text{foot}} &= \begin{cases}
    -h + g_c\sin(\theta_i) & \text{if } \sin(\theta_i) > 0 \\
    -h + g_p\sin(\theta_i) & \text{otherwise}
\end{cases}
\end{align}

\vspace{-0.8em}
\noindent where $d_{step}$ is the maximum step length, $x_{off}$ is the foot offset with respect to the hip, $h$ is the robot height, $g_c$ is the max ground clearance during swing, and $g_p$ is the max ground penetration during stance. A  visualization of the foot trajectory for a set of these parameters is shown in the Pattern Formation block of Figure~\ref{fig:control_diagram}. 

We re-sample $h$, $x_{off}$, $g_c$, and $g_p$ during training so the agent can learn to locomote with varying base heights, foot offsets, swing foot ground clearances, and stance foot ground penetrations. We use the following ranges during training: $h \in[0.18,0.35]$, $x_{off} \in [-0.08,0.03]$, $g_c \in [0.02,0.12]$, $g_p \in [0, 0.015]$.
This is important to vary in order to find the optimal combination, which is unlikely to be the same for each different gait. The agent does not receive any explicit observation of these parameters, and the user can specify each of these parameters during deployment. 

\vspace{-0.3em}
\subsection{Markov Decision Process}
\vspace{-0.2em}
\subsubsection{Action Space} As in CPG-RL~\cite{bellegarda2022cpgrl}, our action space provides an interface for the agent to directly modulate the intrinsic oscillator amplitudes and phases, by learning to modulate $\mu_{i}$ and $\omega_i$ for each leg. This allows the agent to adapt each of these states online in real-time depending on sensory inputs. However, in contrast with our previous work, the strong coupling enforces the relative offsets between different oscillators, meaning the agent is forced to learn parameters to locomote with each particular gait. During training, we resample the coupling matrices randomly among each of the 9 gaits so the agent can learn to locomote with all gaits, as well as transition between different gaits without falling. Our action space can be summarized as $\mathbf{a} = [\bm{\mu}, \bm{\omega}] \in \mathbb{R}^{8}$. The agent selects these parameters at 100 Hz, and we use the following action space ranges during training: $\mu \in [1, 2]$, $\omega \in [0, 8]$ Hz.

\subsubsection{Observation Space}
\label{sec:obs_space}
As in the full observation space of our previous work~\cite{bellegarda2022cpgrl}, our observation space includes velocity commands, the body state (orientation, linear and angular velocities), joint state (positions, velocities), and foot contact booleans. We also include the last action chosen by the policy network and CPG states (i.e.~efference copy of the spinal cord) $\{\bm{r},\bm{\dot{r}},\bm{\theta},\bm{\dot{\theta}}\}$ as feedback to the policy (i.e.~higher centers). Notably, the agent is not directly aware of any coupling matrices (i.e.~gaits), nor mapping parameters $h$, $x_{off}$, $g_c$, $g_p$.

\subsubsection{Reward Function} 
\label{sec:reward_function}

\begin{table}[tpb]
\centering
\vspace{0.6em}
\caption{Reward function terms. $(\cdot)^{*}$ represents a desired command, and $f(x) := \exp{(-\frac{||x||^2}{0.25}) } $. $dt=0.01$ is the control policy time step. 
}
\vspace{-0.2em}
\begin{tabular}{ c c c }
Name & Formula & Weight \\
\hline
Linear velocity tracking $v_{b,x}^{*}$ & $f(v_{b,x}^{*} - v_{b,x})$ & $3 dt$ \\ 
Linear velocity penalty $\bm{v}_{b,yz}$ & $-||\bm{v}_{b,yz}||^2$ & $2dt$ \\
Angular velocity penalty $\bm{\omega}_{b,xyz}$ & $-||\bm{\omega}_{b,xyz}||^2$ & $0.1dt$ \\
Power & $-|\bm{\tau} \cdot \dot{\bm{q}}| $  & $0.001dt$ \\
\hline
\end{tabular} \\
\label{tab:reward}
\vspace{-2em}
\end{table}

Similarly to CPG-RL~\cite{bellegarda2022cpgrl}, our reward function primarily rewards tracking body linear and angular velocity in the base frame. Since in this work we focus on learning gaits during forward locomotion, in addition to forward velocity tracking, we add terms to minimize other undesired base velocities (lateral/vertical oscillations in the base $y$ and $z$ direction, and base roll, pitch, and yaw rates). To minimize energy consumption, we penalize the total power. The terms and respective weights are summarized in Table~\ref{tab:reward}. We emphasize that we do not need to add any reward terms beyond those fully specifying the base motion behavior. Notably, we do not need to specify any `style' rewards to try to enforce any particular gait, base height, foot ground clearance, etc.

\vspace{-0.4em}
\subsection{Training Details}
\vspace{-0.15em}
\label{sec:training_details}

We use Isaac Gym and PhysX as our training environment and physics engine~\cite{isaacgym,rudin2022anymalisaac}, and the Unitree Go1 quadruped~\cite{unitreeGO1}. This framework has high throughput, enabling us to simulate 4096 Go1s in parallel on a single NVIDIA RTX 3090 GPU, which allows us to learn control policies within minutes with the Proximal Policy Optimization (PPO) algorithm~\cite{ppo}. We use the same hyperparameters and neural network architecture as in \cite{bellegarda2022cpgrl}. 

We train on flat terrain, and we reset the environment for an agent if the base or a thigh comes in contact with the ground, or if the episode length reaches 20 seconds. With each reset, we sample new parameters $h$, $g_c$, $g_p$, and $x_{off}$ for mapping the oscillator states to motor commands, allowing the agent to learn continuous locomotion behavior at varying body heights, step heights, and postures. New velocity commands $v_{b,x}^{*}$ are sampled every 5 seconds, and the gait coupling matrix $\bm{\Phi}$ is re-sampled every 3 seconds. As in our previous work, we apply domain randomization on the physical mass properties and coefficient of friction (Table~\ref{table:dyn_rand}). Finally, an external push of up to 0.5 \texttt{m/s} is applied in a random direction to the base every 15 seconds. 

The policy network outputs modulation signals at 100 Hz, and the torques computed from the mapped desired joint positions are updated at 1 kHz. The equations for each of the oscillators (Equations~\ref{eq:salamander_r}-\ref{eq:salamander_theta}) are thus also integrated at 1 kHz. During training we re-sample joint PD controller gains at each environment reset as described in Table~\ref{table:dyn_rand}. 

\vspace{-0.3em}
\section{Experimental Results and Discussion}
\vspace{-0.1em}
\label{sec:result}
In this section we report and discuss results from learning a single controller capable of locomotion with each of the 9 gaits. Sample locomotion policy deployment snapshots are shown in Figure~\ref{fig:intro}, and the reader is encouraged to watch the supplementary video for clear visualizations. 

\begin{figure*}
\vspace{-0.5em}
    \centering
    \includegraphics[width=0.9\linewidth]{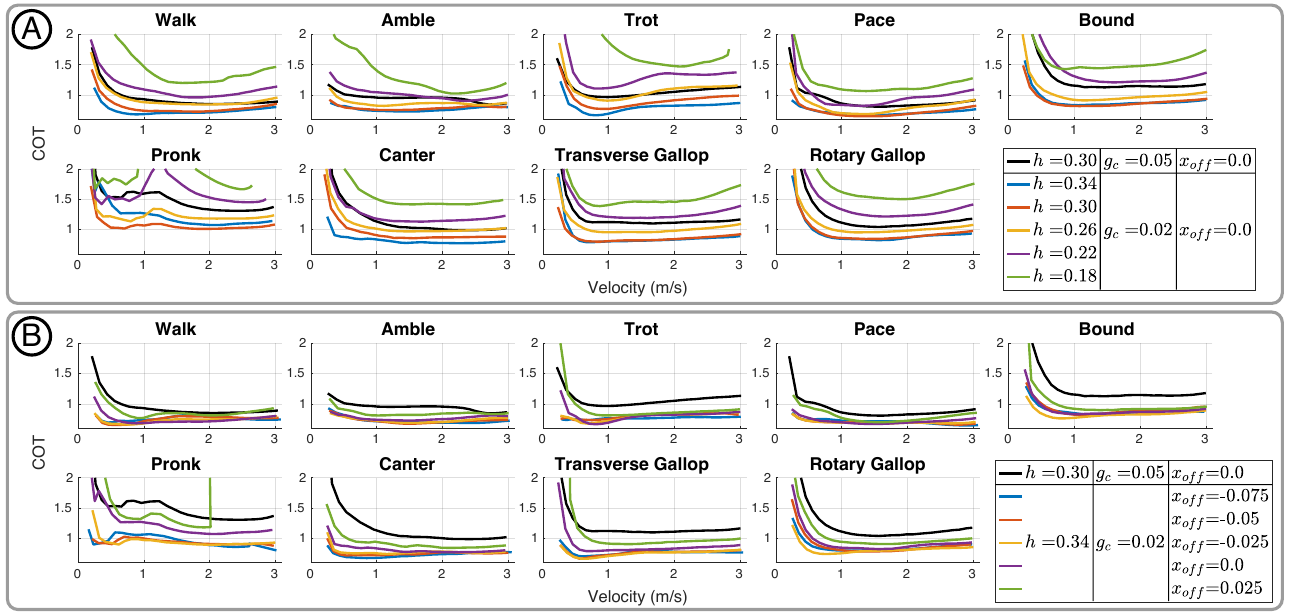}\\
    \vspace{-0.6em}
    \caption{
    Effects on the Cost of Transport for all gaits from modulating nominal (A) body height, and (B) foot offset relative to the hip around which oscillations occur. While the policy is capable of locomotion with all of these gait styles (with notable difficulty for the pronk gait with foot offset 0.025 $m$ in front of the hip above 2 $m/s$), each of these parameters significantly affects the COT, and different combinations of these result in  better energy-efficiency at different velocities, most obviously with respect to the foot offset in (B). 
    }
    \label{fig:cot_examples}
    \vspace{-1.7em}
\end{figure*}

\vspace{-0.3em}
\subsection{Gait Style Parameter Efficiency} 
We first investigate the effects of different gaits and style parameters on the Cost of Transport (COT). After training, we consider the following style parameters (body height $h$, foot ground clearance $g_c$, foot offset $x_{off}$) for each gait: 
\vspace{-0.5em}
\begin{align*}
      h &= \{0.18,\ 0.22,\ 0.26,\ 0.30,\ 0.34\} \\ 
    g_c &= \{0.02,\ 0.05,\ 0.08,\ 0.12\} \\
x_{off} &= \{-0.075,\ -0.05,\ -0.025,\ 0.0,\ 0.025\} 
\end{align*}

\vspace{-0.2em}
\noindent For each of the possible 100 combinations of these three parameters, we command the robot to locomote at 0.3 $m/s$ to 3.0 $m/s$, in increments of 0.1 $m/s$. For each of these 28 velocities, we run the policy for 5 seconds across 100 robots in parallel, and compute the mean Cost of Transport. For the purpose of this data collection, we do not include any noise in the simulation environment. 

\begin{table}[!tpb]
\centering
\caption{Randomized parameters during training and their ranges.}
\vspace{-0.6em}
\begin{tabular}{ c c c c }
Parameter & Lower Bound & Upper Bound & Units \\
\hline 
$v_{b,x}^{*}$    & 0.2 & 3 & \texttt{m/s}\\
\hline
Joint Gain $K_p$ & 30 & 100 & - \\
Joint Gain $K_d$ & 0.5 & 2 & - \\
\hline
Mass (each body link) & 70 & 130 & \%\\
Added base mass & 0  & 5 & $kg$ \\
Coefficient of friction & 0.3 & 1 & -\\
\hline
\end{tabular} \\
\label{table:dyn_rand}
\vspace{-2.4em}
\end{table}

Figure~\ref{fig:cot_examples} shows the effects of varying the three parameters on the Cost of Transport, with respect to baseline parameters seen in previous works such as~\cite{bellegarda2022cpgrl}: $h=0.3$, $g_c=0.05$, $x_{off}=0$, shown by the black line.
Figure~\ref{fig:cot_examples}-A shows the effects of different nominal body heights, from 0.34 $m$ to 0.18 $m$, on the Cost of Transport. As can be expected, a more upright posture generally leads to more efficient locomotion, with lower COT, as less power is needed at the thigh and knee joints to maintain the body height. While almost all gaits have the lowest COT throughout all velocities for the highest, most upright posture, the pronk gait is a notable exception, with the most efficient locomotion for a slightly lower nominal base height parameter, 0.3 $m$. However, many gaits have an almost identical COT curve when locomoting with body height 0.3 $m$ or 0.34 $m$. 
In the video, we also investigate the effects on the COT of changing the nominal swing foot ground clearance from 0.02 $m$ to 0.12 $m$. As can be expected, a lower swing foot ground clearance results in a more efficient gait, as it requires more energy to bring the foot higher off the ground, and this trend is consistent across all gaits. 

Figure~\ref{fig:cot_examples}-B shows the effects of changing the nominal center point around which the oscillations take place with respect to the hip in the $x$ direction range from $-0.075$ $m$ behind the hip, to $0.025$ $m$ in front of the hip. Due to the configuration of the legs and body, the overall Center of Mass (COM) of the robot lies behind the geometric center of the body. Therefore, an offset behind the hips helps to keep the COM more at the center of the foot contacts, thereby (generally) decreasing energy required to remain upright during locomotion. While there is more variability among gaits, and even within a single gait as the locomotion velocity changes, a general observation is that it is more energy-efficient to locomote with the oscillation $x$ center point behind the hips. 

\vspace{-0.4em}
\subsection{Which Gaits and Styles are Most Energy-Efficient?}
\begin{figure}[t]
\vspace{-0.1em}
    \centering
    \includegraphics[width=\linewidth]{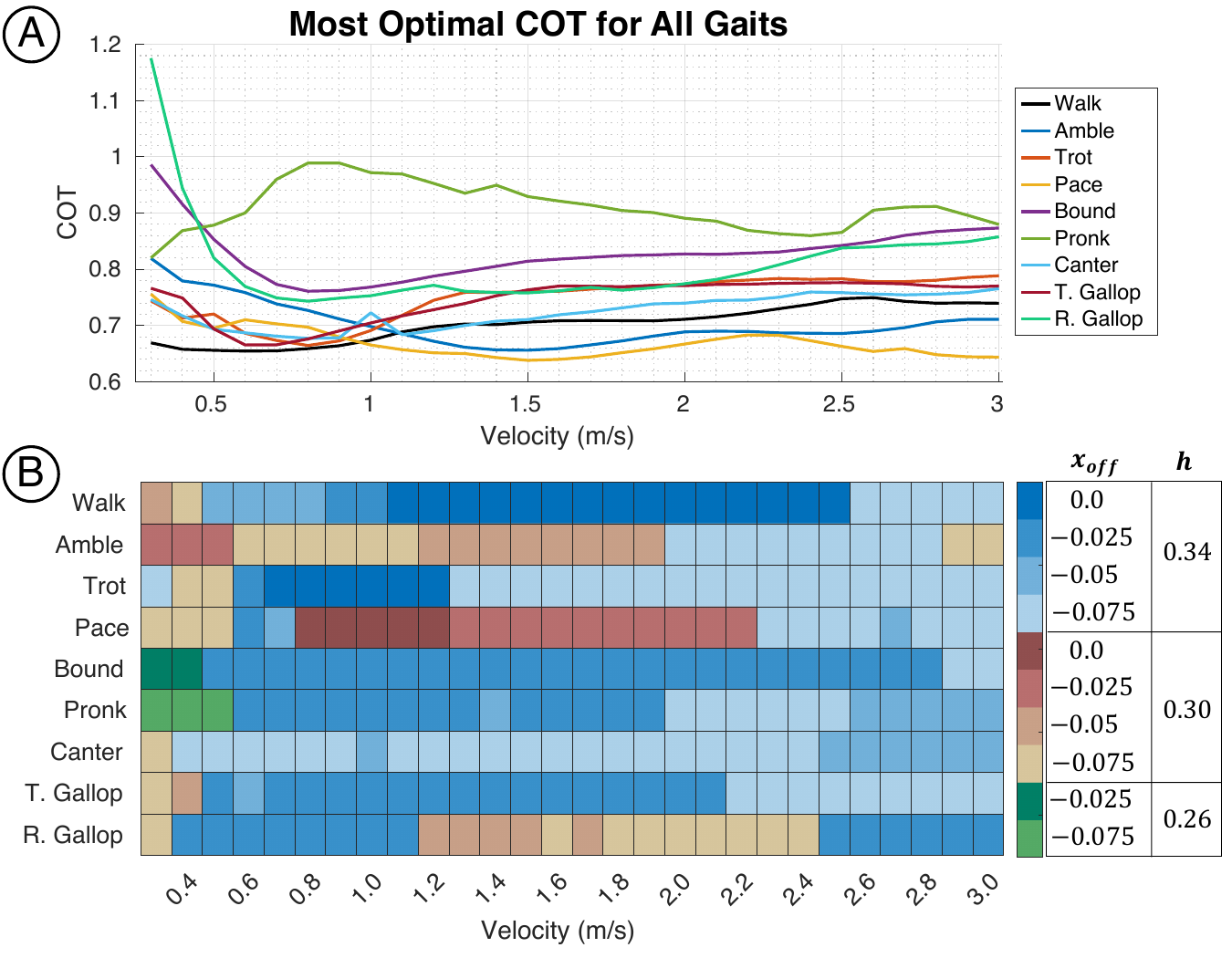}
    \vspace{-1.9em}
    \caption{Most optimal Cost of Transport (COT) for all quadruped gaits, and corresponding gait style parameters for all velocities with our framework. \textbf{(A)}: minimum COT possible for each gait. Walk is most optimal for velocities below 0.9 $m/s$, and pace is most optimal for higher velocities. \textbf{(B)}: corresponding gait style parameters for the minimum COTs in \textbf{(A)}, with varying body heights $h$ and foot offsets $x_{off}$ for all gaits. The lowest COTs here were all achieved with the lowest swing foot ground clearance $g_c = 0.02\ m$. 
    }
    \label{fig:best_COT}
    \vspace{-2em}
\end{figure}

Based on the 100 gait styles considered for each of the 9 gaits, we present the optimal, most energy-efficient locomotion possible for each of the 9 gaits with our framework. Figure~\ref{fig:best_COT} shows the best possible COT for every gait at every velocity at top, and the corresponding gait style parameters at bottom. Between gaits, the walk gait is most energy-efficient from 0.3-0.9 $m/s$, and then the pace gait results in the lowest COT from 0.9-3.0 $m/s$, while the amble gait is second most-efficient in this range. This implies that, contrary to the popular association of modern quadruped robots to dogs or cats, the most efficient locomotion style for our robot makes it closer to a giraffe, camel, elephant, or other pacing and ambling animals.

Notably, as also seen in nature, as locomotion speed changes, different gait styles are more optimal from an energy-efficiency perspective. For all gaits, there does not exist a single set of gait parameters that is optimal at all velocities. However, there are certain important trends. As suggested in the examples in Figure~\ref{fig:cot_examples}, the most optimal parameters at all velocities all use the lowest ground clearance, $0.02$ $m$. For most gaits, especially at higher speeds, the highest nominal body height $0.34$ $m$ gives the most efficient locomotion. However, we see variability in the hip offset, which changes most variably within each gait as a function of speed to subtly improve the COT. 

\begin{figure}[t]
\vspace{-0.5em}
    \centering
    \includegraphics[width=\linewidth,trim={0.4cm 0 1.1cm 0.1cm},clip]{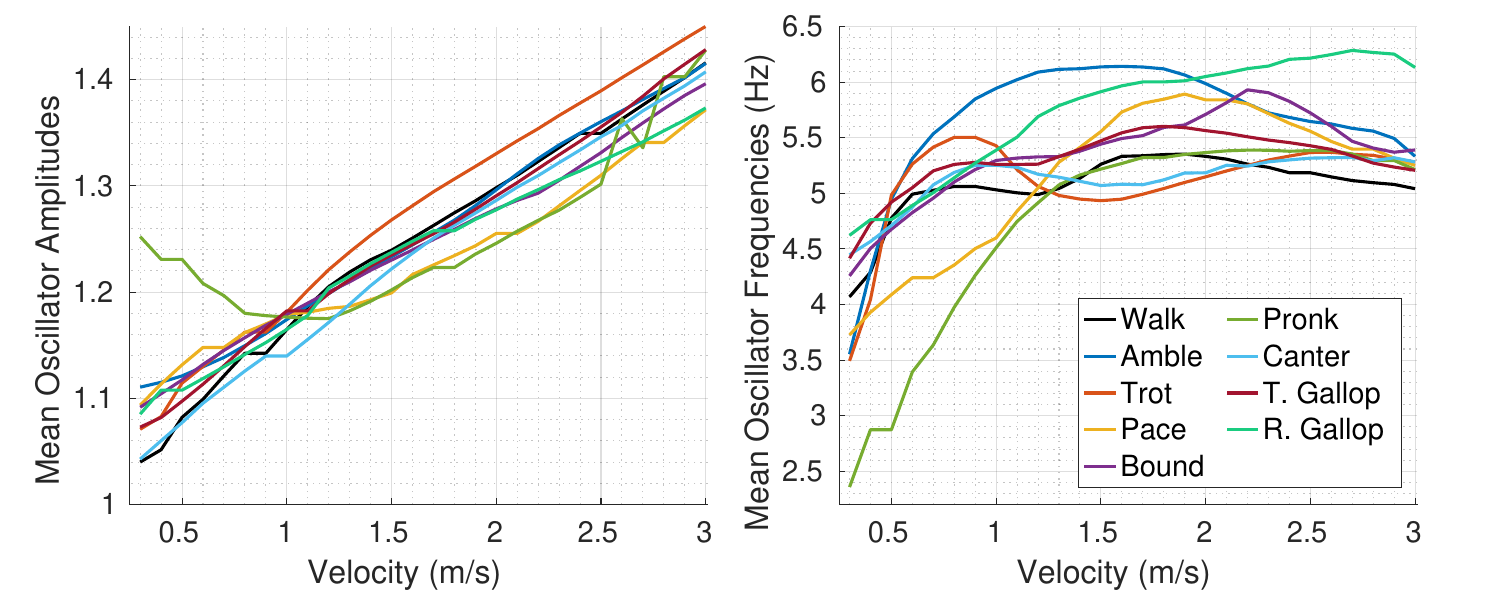} \\
    \vspace{-0.5em}
    \caption{Mean oscillator amplitudes and frequencies for style parameters $h=0.34$, $g_c = 0.02$, $x_{off} = -0.075$, which is optimal for many gaits and at many velocities. Notably, the policy coordinates the amplitudes and frequencies differently for different gaits. 
    }
    \label{fig:cpg_params}
    \vspace{-1.8em}
\end{figure}

One interesting observation for the pronk gait is the reverse curvature with respect to nominal COT plots, which typically have positive parabola-like shapes, with a minimal COT at a particular velocity. For the pronk gait, at low speeds, the policy has learned to modulate the CPG parameters to slowly lean forward, then take a short and fast hop, to more efficiently track the mean desired velocity. This is further shown in Figure~\ref{fig:cpg_params} by the mean CPG amplitude and frequency selected by the policy to locomote as a function of speed, which changes for each gait and style. The amplitude and frequency of the oscillators directly correspond to mean step length and mean step frequency. Interestingly, we see several inflection points for several gaits, where the strategy changes for increasing locomotion speed. This is most obvious for the pace and amble gaits, which actually decrease the step frequency at higher speeds, while instead locomoting faster by increasing the step length. The policy has learned this different coordination among gaits in order to maximize the returns from our reward function, and without explicit knowledge of the coupling parameters or gait style, but which it can indirectly deduce through observing the CPG and joint states.  
Thus, our framework can also be used as a tool to evaluate different locomotion strategies given a particular gait. 

\vspace{-0.3em}
\subsection{Is Energy-Efficiency the Most Important Gait Metric?}
\vspace{-0.15em}
While energy-efficiency is often the most popular explanation for different gait styles at different speeds, why do walk and pace gaits not naturally emerge when training locomotion policies with end-to-end deep reinforcement learning for quadruped robots? Indeed, most recent results that learn robot locomotion on flat or rough terrain show a trot gait both at high and low velocities~\cite{ji2022concurrent,margolis2022rapid,miki2022learning,rudin2022anymalisaac}. However, in addition to velocity tracking, typically the reward function includes many terms related to base stability to keep smooth motions, minimize joint accelerations, limit large forces, keep the feet underneath the hips, and promote foot air time to encourage a ``natural'' gait. In fact, from Figure~\ref{fig:cot_examples}, we can observe that the trot gait is best at low velocities from the COT point of view \textit{if} the foot oscillation offset is underneath the hip for all gaits ($x_{off}=0$).  Therefore, it can be expected that the COT is not the only (or even primary) objective for most existing work, where velocity tracking and gait robustness/stability are important and considerable factors. 

\begin{figure}[!t]
    \centering
    \includegraphics[width=0.49\linewidth,trim={0 0 1.15cm 0.5cm},clip]{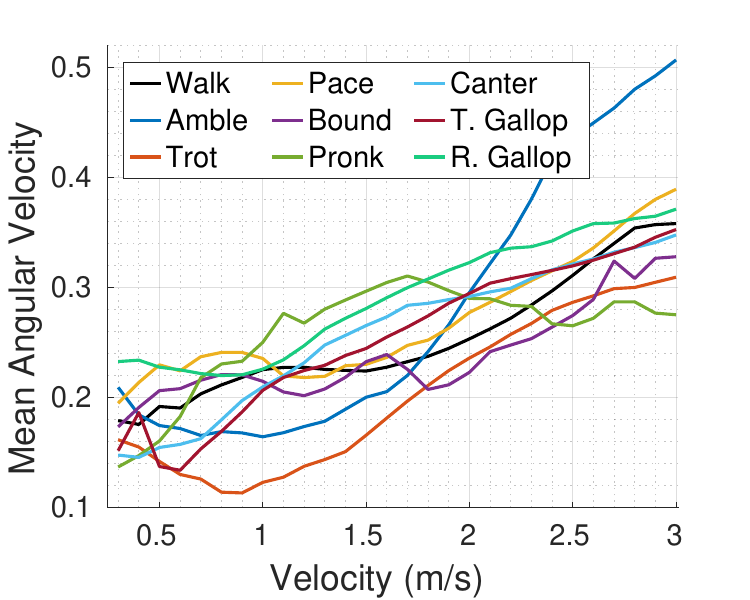}
    \includegraphics[width=0.49\linewidth,trim={0 0 1.15cm 0.5cm},clip]{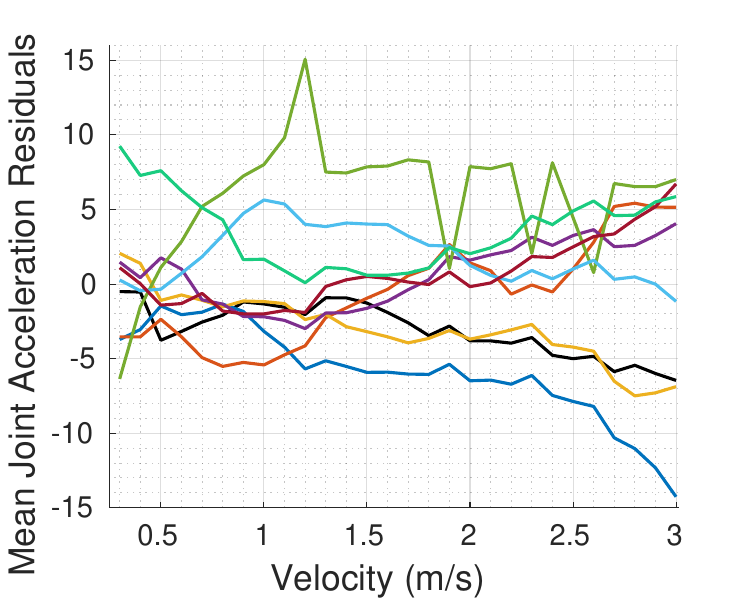} \\
    \vspace{-0.3em}
    \caption{Minimum mean angular velocity of the base (left), and joint acceleration (right), across all gaits and gait style parameters for all velocities. At right, the residuals with respect to the mean joint acceleration for all gaits at that same particular  velocity is shown for clarity purposes, with negative values indicating gaits with lower joint acceleration. 
    }
    \label{fig:best_angvel_jerk}
    \vspace{-2em}
\end{figure}

In Figure~\ref{fig:best_angvel_jerk}, we evaluate our policy for each gait with all possible styles with respect to two different metrics: mean angular velocity, and mean joint acceleration. The mean angular velocity is the average of the absolute sum of all three angular velocities of the body during locomotion at a particular forward velocity. To make the distinction more obvious, we plot the joint acceleration residuals with respect to the mean joint acceleration among all gaits at a particular velocity. It is important to note that the gait styles that minimize mean angular velocity or mean joint acceleration are not the same, nor are they the same as those that minimize the COT. In fact, for both metrics, they vary much more frequently and in larger ranges depending on the speed. From the plots, we can observe that a gait with the lowest COT at a particular velocity may not be optimal with respect to other metrics. For example, at low velocities, the trot gait has the lowest mean angular velocity, indicating a stable base, while having only a slightly higher COT w.r.t. the walk gait. At higher velocities, the bound gait has lower mean angular velocity (which typically does not have much roll or yaw), while at the highest velocities, the pronk gait is best (which typically locomotes with a flat base throughout the motion). While the amble gait has the highest mean angular velocity at high speeds, we see that it has the lowest mean joint acceleration, and has the second best COT (Figure~\ref{fig:best_COT}). Therefore, depending on the importance a user gives to different metrics, this directly changes the ``optimal'' gait. For example, if tasked with carrying a fragile load at low speeds, the trot gait might be best as it has the best base stability.

\vspace{-0.4em}
\subsection{Qualitative Evaluation}
\vspace{-0.2em}
\subsubsection{All Gaits and Transitions} In the video, we show both simulation and hardware results from deploying our single policy to accomplish all 9 gaits, with a variety of styles (i.e.~varying body height, swing foot height, and foot offsets). Notably, we are able to transition between any gaits at any velocity, just by re-sampling the coupling matrix, without any explicit input to the policy network. We can also modulate the gait style parameters within the same gait, or switch these along with the gait, by changing the pattern formation parameters. 

\subsubsection{Novel Gaits} We test the robustness and ability of the policy to produce novel gaits not seen during training. Since we had already trained for all quadruped gaits, we develop several artificial gaits (not found in nature) by creating new coupling matrices. For example, we test new gaits where three limbs are coupled in phase, and one is out of phase with a $\pi$ offset. The out of phase leg can be either a front or rear leg, and the video shows that although these gaits and observations were never seen during training, we can still effectively locomote with such new gaits at test time in hardware experiments. We also test different timing patterns for several gaits, for example modifying the trot or bound with different timing to create more uneven gaits, which still result in successful locomotion and velocity tracking. 

\subsubsection{Leg Failure Robustness} Lastly, we show that our single policy is robust among different gaits to disabling either one or two legs. We design an experiment where we transition from trot, to pace, to bound, to pronk; and then repeat the experiment with disabling (locking) one or two rear legs, so that they stay fixed in a nominal extension position. The robot does not fall down and continues to locomote during these experiments, despite never having encountered any of these situations during training.

\vspace{-0.3em}
\section{Conclusion}
\label{sec:conclusion}
In this work, we have presented a framework for realizing all quadruped gaits and transitions between these gaits, all of which can be done with a variety of styles (body heights, swing foot ground clearances, and foot offsets). We trained a single locomotion policy to modulate the amplitudes and frequencies of a network of coupled oscillators which can enforce a variety of gaits, the output of which is mapped through a pattern formation layer to produce locomotion with different styles. Notably, we used a simple reward function, and did not need any expert demonstrations, providing a simple and effective framework for realizing all gaits. We found the policy capable of producing various novel and artificial (not found in nature) gaits not seen during training, and observed robustness to any gait transitions with any styles, and to one or two leg failures.

We investigated the energy-efficiency of each gait as a function of locomotion velocity and style, and found the most efficient gaits for our quadruped with our framework is the walk gait at low velocities, and the pace gait at higher velocities. This result implies, based on the morphology and mechanics of the robot and their effects on the Cost of Transport, that the robot may be closer to a pacing or ambling mammal such as a camel or elephant, rather than the more popular association with a dog or cat. However, contrary to the ``best gait'' from an energy-efficiency point of view, other gaits like trot result in a more stable base at low velocities, and the amble gait results in lower joint acceleration at higher velocities, warranting further investigation for different robot morphologies. 

\bibliographystyle{IEEEtran}
\bibliography{refs}

\end{document}